\documentclass[review]{elsarticle}

\usepackage{adjustbox}
\usepackage{algorithm}
\usepackage{algorithmic}
\usepackage{amsmath,amssymb,amsfonts,amsthm}
\usepackage{array}
\usepackage{blkarray}
\usepackage{bm}
\usepackage{booktabs}
\usepackage{diagbox}
\usepackage{dsfont}
\usepackage[inline]{enumitem}
\usepackage{graphicx}
\usepackage{hyperref}
\usepackage{makecell}
\usepackage{mathtools}
\usepackage{multirow}
\usepackage{nth}
\usepackage{pdflscape}
\usepackage{siunitx}
\usepackage{soul}
\usepackage{stmaryrd}
\usepackage{subcaption}
\usepackage{tabularx}
\usepackage{textcomp}
\usepackage{xcolor}
\usepackage{ulem}

\usepackage[mode=buildnew]{standalone}
\usepackage{pgfplots}
\usepackage{tikz}
\usetikzlibrary{patterns}

\newcommand{\norm}[1]{\left\lVert#1\right\rVert}

\newcommand{\nint}[1]{\llbracket#1\rrbracket}

\def\*#1{\mathbf{#1}}

\newcommand{\conv}[1]{\smash{\overset{\scriptscriptstyle\smile}{#1}}}
\newcommand{\conc}[1]{\smash{\overset{\scriptscriptstyle\frown}{#1}}}
\def\defequal{\stackrel{\mbox{\footnotesize def}}{=}}

\newcommand{\bHtilde}{\boldsymbol{\tilde{\mathbf{H}}}}
\newcommand{\bVtilde}{\boldsymbol{\tilde{\mathbf{V}}}}
\newcommand{\bWtilde}{\boldsymbol{\tilde{\mathbf{W}}}}
\newcommand{\bHcheck}{\boldsymbol{\check{\mathbf{H}}}}
\newcommand{\bWcheck}{\boldsymbol{\check{\mathbf{W}}}}

\newcommand{\EE}{\mathbb{E}}

\newcommand{\RR}{\mathbb{R}}

\DeclareMathOperator*{\argmin}{arg\,min}

\SetSymbolFont{stmry}{bold}{U}{stmry}{m}{n}

\begin{document}

% =========================================================================== %

\begin{frontmatter}

  \title{Joint Majorization-Minimization for Nonnegative Matrix
  Factorization with the $\beta$-divergence}
  
  \author[AMU]{Arthur Marmin\corref{my_corres_author}}
  \cortext[my_corres_author]{Corresponding author}
  \ead{arthur.marmin@univ-amu.fr}
  
  \author[INP]{Jos{\'e}~Henrique~de~Morais~Goulart}
  \ead{henrique.goulart@irit.fr}
  
  \author[IRIT]{C{\'e}dric~F{\'e}votte}
  \ead{cedric.fevotte@irit.fr}

  \address[AMU]{Aix Marseille Universit{\'e}, CNRS, I2M, UMR 7373\\
    Marseille, France}
  \address[IRIT]{IRIT, Universit{\'e}
    de Toulouse, CNRS,\\
    Toulouse, France}
  \address[INP]{IRIT, Universit{\'e} de Toulouse,
    Toulouse INP,\\
    Toulouse, France}
  
  % -------------------------------------------------------------------------- %
  
  \begin{abstract}
    This article proposes new multiplicative updates for nonnegative matrix
    factorization (NMF) with the $\beta$-divergence objective function.
    Our new updates are derived from a joint majorization-minimization (MM)
    scheme, in which an auxiliary function (a tight upper bound of the objective
    function) is built for the two factors jointly and minimized at each
    iteration.
    This is in contrast with the classic approach in which a majorizer is derived
    for each factor separately.
    Like that classic approach, our joint MM algorithm also results in
    multiplicative updates that are simple to implement.
    They however yield a significant drop of computation time (for equally good
    solutions), in particular for some $\beta$-divergences of important applicative
    interest, such as the quadratic loss and the Kullback-Leibler or
    Itakura-Saito divergences.
    We report experimental results using diverse datasets: face images, an audio
    spectrogram, hyperspectral data and song play counts.
    Depending on the value of $\beta$ and on the dataset, our joint MM approach can
    yield CPU time reductions from about $13\%$ to $86\%$ in comparison to the
    classic alternating scheme.
  \end{abstract}
  
  % -------------------------------------------------------------------------- %
  

  \nonumnote{This work is supported by the European Research Council
    (ERC FACTORY-CoG-6681839), the French Agence Nationale de la Recherche
    (ANITI, ANR-19-P3IA-0004) and the National Research Foundation, Prime
    Minister’s Office, Singapore under its Campus for Research Excellence and
    Technological Enterprise (CREATE) programme.}

\end{frontmatter}

% =========================================================================== %

\section{Introduction}

Nonnegative matrix factorization (NMF) aims at factorizing a matrix with
nonnegative entries into the product of two nonnegative matrices.
It has found many applications in various domains which include feature
extraction in image processing and text
mining~\cite{Lee_D_1999_j-nature_learning_ponmf}, audio source
separation~\cite{Smaragdis_P_2014_j-ieee-sig-proc-mag_static_dssunmfuv},
blind unmixing in hyperspectral
imaging~\cite{Berry_W_2007_j-comput-stat-data-anal_algorithms_aanmf,
  BioucasDias_J_2012_j-ieee-j-sel-top-appl-earth-obs-rem-sens_hyperspectral_uogssrba},
and user recommendation~\cite{Hu_Y_2008_p-ieee-icdm_collaborative_fifd}.
See~\cite{Cichoki_A_2009_book_nonnegative_mtf,
  Fu_X_2019_j-ieee-sig-proc-mag_nonnegative_mfsdaiaa,
  Gillis_N_2020_book_nonnegative_mf} for overview papers and books about NMF\@.

Each application gives different interpretations to the factor matrices but the
first factor is often considered as a dictionary of recurring patterns while the
second one describes how the data samples are expanded onto the dictionary
(activation matrix).
The nonnegativity constraint only allows additive combination of the dictionary
elements, yielding meaningful additive and sparse representations of the data.

Computing an NMF generally consists in minimizing a well-chosen objective
function under nonnegativity constraints.
A popular choice of objective function is the $\beta$-divergence, which is a
continuous family of measures of fit parameterized by a single parameter $\beta$
that encompasses the Kullback-Leibler (KL) or Itakura-Saito (IS) divergences as
well as the common squared Euclidean
distance~\cite{Cichocki_A_2011_j-entropy_generalized_abdtarnmf}.
In the latter cases, the $\beta$-divergence is a log-likelihood in disguise for
Poisson, multiplicative Gamma and additive Gaussian noise models, respectively.

The classic approach to NMF, and to NMF with the $\beta$-divergence in particular,
consists in optimizing the two factors alternately, i.e., using two-block
coordinate descent.
Each of the two factors is then updated using Majorization-Minimization (MM), as
described in~\cite{Lee_D_2001_adv_neur_proc_inf_algorithms_nmf,
  Kompass_R_2007_j-neural-comput_generalized_dmnmf,
  Nakano_M_2010_p-ieee-mlsp_convergence_gmanmfbd,
  Fevotte_C_2011_j-neural-comput_algorithms_nmfbd,
  Yang_Z_2011_j-ieee-trans-nnet_unified_dmalqnmf}: given one of the factors, a
tight upper bound of the objective function is constructed and minimized with
respect to (w.r.t) the other factor.
This results in multiplicative updates that are simple to implement (with no
hyperparameter to tune), have linear complexity per iteration, and that
automatically preserve nonnegativity given positive initializations.
By construction, MM ensures monotonicity of the objective function
(non-increasing values), see~\cite{Lange_K_2016_book_mm_oa,
  Sun_Y_2017_j-ieee-trans-sig-proc_majorization_maspcml} for tutorials about
MM\@.

Thanks to its simplicity, the block-descent approach is dominant in matrix
factorization and dictionary learning (using sometimes other block partitions
such as columns or rows~\cite{Cichoki_A_2009_book_nonnegative_mtf,
  Gillis_N_2020_book_nonnegative_mf}) and very few works have addressed joint
(``all-at-once'') optimization of the factors, though the latter approach may be
more efficient.
A notable exception in dictionary learning (real-valued factors, sparse
activations, quadratic loss) is~\cite{Rakotomamonjy_A_2013_j-ieee-trans-sig-proc_direct_odlp}.
In this work, the author employs an elegant non-convex proximal splitting
strategy and shows that the joint approach is significantly faster than
alternating methods without altering the quality of the obtained solution.
In a similar spirit to~\cite{Rakotomamonjy_A_2013_j-ieee-trans-sig-proc_direct_odlp},~\cite{Mukkamala_M_2019_p-nips_beyond_aumfibpga}
leverages the theoretical framework of~\cite{Bolte_J_2018_j-siam-j-optim_first_ombclgcaqip}
to address matrix factorization (including NMF) with non-alternating updates.
Their work relies on the generalization of Lipschitz-continuity of the gradient
(which does not hold jointly for both factors) to adaptive smoothness~\cite{Bolte_J_2018_j-siam-j-optim_first_ombclgcaqip}.
Yet, their results only apply to the quadratic loss and Newton-like acceleration
is crucial to obtain competitive results. Using tools from dynamical systems,
the authors in~\cite{Panageas_I_2020_PP_convergence_sosnmfpc} have derived a
novel form of multiplicative updates which can run concurrently for each factor
matrix at a given iteration.
They have the additional benefit of ensuring the convergence to a local
minimum instead of a mere critical point.
However, their results are again limited to the quadratic loss.
This also applies to~\cite{Marumo_N_2023_j-comput-optim-appl_majorization_mlmmcnls}
where a Levenberg-Marquardt joint optimization method is described for NMF with
the quadratic loss.
In~\cite{Vandecappelle_M_2020_j-ieee-trans-sig-proc_second_omfcpdnlsc}, the
authors propose a joint second-order Newton-like algorithm for nonnegative
canonical tensor decomposition with the $\beta$-divergence, which takes NMF as a
special case.
Their approach relies on approximations of the Hessian matrix, sometimes based
on heuristics, and fails to provide an algorithm that universally works for
every value of $\beta$.

Inspired by these works, we propose a joint MM approach to $\beta$-NMF and compare
it to the classic block MM strategy.
Our joint MM relies on a tight majorization of the objective function with
respect to all its variables instead of using blocks of fixed variables.
Iterative minimization of this joint upper bound results in new multiplicative
updates that are simple but potentially more efficient variants of the classical
multiplicative updates.
This is particularly true when considering the quadratic loss and
the KL or IS divergences for which further simplifications occur.
We show in these cases that our update rules decrease the computation cost per
iteration.
It turns out that our joint upper bound coincides with the one derived
in~\cite{Takahashi_N_2018_j-comput-optim-appl_unified_ugcamurnmf}.
The latter is however employed for a different purpose, namely the convergence
analysis of classic block MM for NMF, and not to design a new algorithm like we
do (more details will be given in Section~\ref{ssec:discussion_jmm}).

Our methodological results are supported by extensive simulations using datasets
with different sizes arising from various applications in which NMF had a
significant impact (face images, audio spectrogram, hyperspectral images, song
play-counts).
In most scenarii, the proposed joint MM approach leads to a reduction of the
computing time ranging from $13\%$ to $86\%$ without incurring any loss in the
precision of the approximation.

The article is organized as follows: Section~\ref{sec:prelim} states the NMF
optimization problem and summarizes the classic block MM approach.
Section~\ref{sec:jmm} first presents our proposed joint MM method for NMF and
derives the new multiplicative updates.
It then compares the joint and the block MM methods in term of computational
complexity, and discusses the benefit of the joint approach.
Comparative numerical simulations are presented in Section~\ref{sec:simul} and
validate the efficiency of our approach.
Finally, Section~\ref{sec:concl} draws conclusions.

\paragraph*{Notation} $\RR_{+}$ is the set of nonnegative real numbers, and
$\nint{1,N}$ is the set of integers from $1$ to $N$.
Bold upper case letters denote matrices, bold lower case letters denote vectors,
and lower case letters denote scalars.
The notation ${[\*M]}_{ij}$ and $m_{ij}$ both stand for the element of $\*M$
located at the $i^{\text{th}}$ row and the $j^{\text{th}}$ column.
For a matrix $\*M$, the notation $\*M \ge 0$ denotes entry-wise nonnegativity.

% ============================================================================ %

\section{Preliminaries}
\label{sec:prelim}

\subsection{Nonnegative matrix factorization}

We aim at factorizing a $F \times N$ nonnegative matrix $\*V$ into the product
$\*W\*H$ of two nonnegative factor matrices of sizes $F \times K$ and $K \times N$,
respectively.
The rank value $K$ is often chosen such that $FK + KN \ll FN$,
leading to a low-rank approximation of $\*V$.
Given a desired value for the rank $K$, the factor matrices are obtained by
solving the following optimization problem
\begin{equation}
  \label{eq:nmf}
  \min_{\*W, \*H \ge 0 } D_{\beta}(\*V \mid \*W\*H)  \, ,
\end{equation}
where $D_{\beta}$ is a separable objective function defined by
\begin{equation}
  \label{eq:obj_fct}
  D_{\beta}(\*V \mid \*W\*H) = \sum_{f=1}^{F}\sum_{n=1}^{N} d_\beta(v_{fn}\mid{[\*W\*H]}_{fn}) \, .
\end{equation}
Our measure of fit $d_{\beta}$ is the
$\beta$-divergence~\cite{Cichocki_A_2011_j-entropy_generalized_abdtarnmf} given by
\begin{equation}
  \label{eq:beta-div}
  d_{\beta}(x \mid y) =
  \begin{cases*}
    x\log \frac{x}{y} - x + y          & if $\beta=1$ \\
    \frac{x}{y} - \log \frac{x}{y} - 1 & if $\beta=0$ \\
    \frac{x^{\beta}}{\beta(\beta-1)}+\frac{y^{\beta}}{\beta} - \frac{x y^{\beta-1}}{\beta-1}
    & otherwise.
  \end{cases*}
\end{equation}
The value of $\beta$ is chosen according to the application context and the noise
assumptions on $\*V$~\cite{Fevotte_C_2011_j-neural-comput_algorithms_nmfbd}.
The IS divergence, KL divergence and quadratic loss are obtained
for $\beta= 0,1,2$, respectively.

% ---------------------------------------------------------------------------- %

\subsection{Classic multiplicative updates}
\label{ssec:bmm}

The classic method to solve Problem~\eqref{eq:nmf} consists in a two-block
coordinate descent approach where each block is handled with MM\@.
Namely, it alternately minimizes $(\*W,\*H) \mapsto D_{\beta}(\*V|\*W\*H)$ in $\*W$ and in
$\*H$ using MM\@.
The MM method is a two-step iterative optimization
scheme~\cite{Lange_K_2016_book_mm_oa,
  Sun_Y_2017_j-ieee-trans-sig-proc_majorization_maspcml}.
At each iteration, the first step consists in building a local auxiliary
function $G$ that is minimized in the second step.
The auxiliary function has to be a tight majorizer of the original objective
function $\phi:\EE\mapsto\RR$ at the current iterate $\tilde{\*x}$, where $\EE$ is the
domain of $G$ and $\phi$.
More precisely, it has to satisfy the following two properties:
\begin{align*}
  \left(\forall\*x\in\EE\right) \quad
  G(\*x \mid \tilde{\*x}) &\geq \phi(\*x) \\
  G(\tilde{\*x} \mid \tilde{\*x}) &= \phi(\tilde{\*x}) \, .
\end{align*}
These properties ensure that any iterate $\*x$ that decreases the value of $G$
also decreases the value of $\phi$.
Indeed, for a given $\tilde{\*x}$, if we find $\*x$ such that
$G(\*x|\tilde{\*x}) \leq G(\tilde{\*x}|\tilde{\*x})$, then the tight majorization
properties induce the following descent lemma
\begin{equation}
  \label{eq:desc_lemm}
  \phi(\*x) \leq G(\*x \mid \tilde{\*x}) \leq G(\tilde{\*x} \mid \tilde{\*x}) = \phi(\tilde{\*x})
  \, .
\end{equation}
Note that even if $G$ is not minimized but only decreased in value, the descent
property still holds.

The previous MM scheme can be applied alternately to the minimization of the two
functions $\*W \mapsto D_{\beta}(\*V|\*W\*H)$ and $\*H \mapsto D_{\beta}(\*V|\*W\*H)$.
These two functions are the sum of concave, convex, and constant terms.
A convex auxiliary function can then be easily derived by using Jensen's
inequality for the convex term and the tangent inequality for the concave term,
see~\cite{Fevotte_C_2011_j-neural-comput_algorithms_nmfbd} and the next section.
This yields the following multiplicative updates
\begin{equation}
  \label{eq:alt_update}
  \begin{aligned}
    \*W &\; \longleftarrow \; \*W.{\left(\frac{\left({(\*W\*H)}^{.(\beta-2)}.\*V\right)\*H^{\top}}
      {\left({(\*W\*H)}^{.(\beta-1)}\right)\*H^{\top}}\right)}^{.\gamma(\beta)} \\
    \*H &\; \longleftarrow \; \*H.{\left(\frac{\*W^{\top}\left({(\*W\*H)}^{.(\beta-2)}.\*V\right)}
      {\*W^{\top}\left({(\*W\*H)}^{.(\beta-1)}\right)}\right)}^{.\gamma(\beta)}
    \, ,
  \end{aligned}
\end{equation}
where $.$ and $/$ are the entry-wise multiplication and division, respectively,
and $\gamma(\beta)$ is a scalar defined as
\begin{equation}
  \label{eq:def_gamma}
  \gamma(\beta) =
  \begin{cases*}
    \frac{1}{2-\beta} & if $\beta \in ]-\infty,1[$ \\
                1 & if $\beta\in[1,2]$ \\
    \frac{1}{\beta-1} & if $\beta\in]2,+\infty[$
  \end{cases*} \, .
\end{equation}
Note that by construction, the matrices $\*W$ and $\*H$ resulting from the
update~\eqref{eq:alt_update} contain only positive coefficients if the input
matrices $\*V$, $\*W$, and $\*H$ are all positive.
The nonnegativity of the iterates is thus preserved by the multiplicative
structure of the updates given positive initializations.

Note that strict MM dictates that the individual updates of $\*W$ and $\*H$
given in~\eqref{eq:alt_update} shall be applied several times to fully minimize
the partial functions $\*W \mapsto D_{\beta}(\*V|\*W\*H)$ and $\*H \mapsto D_{\beta}(\*V|\*W\*H)$.
This leads to Algorithm~\ref{algo:bmm_sketch}, that we refer to as Block MM
(BMM).
Note that in common NMF practice, only one sub-iteration is used
($L_{W}=L_{H}=1$), which still results in a descent algorithm thanks to the
descent lemma~\eqref{eq:desc_lemm}.

\begin{algorithm}[t]
  \begin{algorithmic}[1]
    \renewcommand{\algorithmicrequire}{\textbf{Input:}}
    \renewcommand{\algorithmicensure}{\textbf{Output:}}
    \REQUIRE{Nonnegative matrix $\*V$ and initialization}
    $(\*W_{\text{init}},\*H_{\text{init}})$
    \ENSURE{Nonnegative matrices $\*W$ and $\*H$ such that $\*V \approx \*W\*H$}
    \STATE{Initialize $i$ to $1$}
    \STATE{Initialize $(\bWcheck_{i},\bHcheck_{i})$ to
      $(\*W_{\text{init}},\*H_{\text{init}})$}
    \REPEAT{}
    \STATE{Initialize $\*W_{1}$ to $\bWcheck_{i}$}
    \FOR{$l=1 \dots L_{W}$} 
    \STATE{Update $\*W$ using~\eqref{eq:alt_update}
      \[
      \*W_{l+1} \longleftarrow
      \*W_{l}.{\left(\frac{\left({(\*W_{l}\bHcheck_{i})}^{.(\beta-2)}.\*V\right)\bHcheck_{i}^{\top}}
        {\left({(\*W_{l}\bHcheck_{i})}^{.(\beta-1)}\right)\bHcheck_{i}^{\top}}\right)}^{.\gamma(\beta)}
      \]}
    \ENDFOR{}
    \STATE{$\bWcheck_{i+1} \leftarrow \*W_{L_{W}+1}$}
    \newline
    \STATE{Initialize $\*H_{1}$ to $\bHcheck_{i}$}
    \FOR{$l=1 \dots L_{H}$}
    \STATE{Update $\*H$ using~\eqref{eq:alt_update}
      \[
      \*H_{l+1} \longleftarrow
      \*H_{l}.{\left(\frac{\bWcheck_{i+1}^{\top}\left({(\bWcheck_{i+1}\*H_{l})}^{.(\beta-2)}.\*V\right)}
      {\bWcheck_{i+1}^{\top}\left({(\bWcheck_{i+1}\*H_{l})}^{.(\beta-1)}\right)}\right)}^{.\gamma(\beta)}
      \]}
    \ENDFOR{} \\
    \STATE{$\bHcheck_{i+1} \leftarrow \*H_{L_{H}+1}$}
    \newline
    \STATE{Increment $i$}
    \UNTIL{Convergence}
    \RETURN{$(\bWcheck_{i},\bHcheck_{i})$}
  \end{algorithmic}
  \caption{BMM~\label{algo:bmm_sketch}}
\end{algorithm}

% ============================================================================ %

\section{Joint Majorization-Minimization}
\label{sec:jmm}

In contrast with the classic alternating scheme presented in
Section~\ref{ssec:bmm}, we develop in this section a joint MM (JMM) approach for
solving Problem~\eqref{eq:nmf}.

% ---------------------------------------------------------------------------- %

\subsection{Construction of the auxiliary function}

In order to apply the MM scheme, we start by looking for a suitable auxiliary
function $G: \RR_{+}^{F \times K}\times\RR_{+}^{K \times N} \rightarrow \RR_{+}$ for the function
$(\*W,\*H) \mapsto D_{\beta}(\*V|\*W\*H)$.
We observe in~\eqref{eq:obj_fct} that $D_{\beta}$ is a sum of $FN$ $\beta$-divergences
between scalars.
Our approach is to construct an auxiliary function for each summand.
Following~\cite{Fevotte_C_2011_j-neural-comput_algorithms_nmfbd}, the
$\beta$-divergence $d_{\beta}$, taken as a function of its second argument, can be
decomposed into the sum of a convex term $\conv{d}_{\beta}$, a concave term
$\conc{d}_{\beta}$, and a constant term $\bar{d}_{\beta}$.
The definitions of these three terms for the different values of $\beta$ are given
in Table~\ref{table:elmt_beta-div}. 

\begin{table}[t]
  \caption{Decomposition of $d_{\beta}$ for the different values of $\beta$.}
  \begin{center}
    \setlength\tabcolsep{5.5pt}
    \resizebox{\columnwidth}{!}{\begin{tabular}{cccc}
      \toprule
      $\beta$ & $\conv{d}_{\beta}(v_{fn} \mid {[\*W\*H]}_{fn})$
          & $\conc{d}_{\beta}(v_{fn} \mid {[\*W\*H]}_{fn})$
          & $\bar{d}_{\beta}(v_{fn})$ \\
      \midrule
      \midrule
      $]-\infty,1[\setminus\{0\}$ & $\frac{-v_{fn}}{\beta-1}{[\*W\*H]}_{fn}^{\beta-1}$
                   & $\frac{{[\*W\*H]}_{fn}^{\beta}}{\beta}$
                   & $\frac{v_{fn}^{\beta}}{\beta(\beta-1)}$ \\
      \midrule
      $0$          & $\frac{v_{fn}}{{[\*W\*H]}_{fn}}$
                   & $\log {[\*W\*H]}_{fn}$
                   & $-(\log v_{fn} + 1)$ \\
      \midrule
      $[1,2]$      & $d_{\beta}(v_{fn} \mid {[\*W\*H]}_{fn})$ & $0$ & $0$ \\
      \midrule
      $]2,+\infty[$     & $\frac{{[\*W\*H]}_{fn}^{\beta}}{\beta}$
                   & $\frac{-v_{fn}}{\beta-1}{[\*W\*H]}_{fn}^{\beta-1}$
                   & $\frac{v_{fn}^{\beta}}{\beta(\beta-1)}$ \\
      \bottomrule
    \end{tabular}}
  \end{center}
  \label{table:elmt_beta-div}
\end{table}

Next, we majorize the convex and concave terms of $d_{\beta}$ separately.
The methodology
follows~\cite{Fevotte_C_2011_j-neural-comput_algorithms_nmfbd}, except that none
of the two factors $\*W$ or $\*H$ is treated as a fixed variable.
The convex term $\conv{d}_{\beta}$ is majorized using Jensen's inequality
\begin{align}
  \conv{d}_{\beta}(v_{fn} \mid {[\*W\*H]}_{fn})
  &= \conv{d}_{\beta}\left(v_{fn} \;\middle|\; \sum_{k=1}^{K}w_{fk}h_{kn}\right)
  \nonumber \\
  &= \conv{d}_{\beta}\left(v_{fn} \;\middle|\;
  \sum_{k=1}^{K}\tilde{\lambda}_{fnk}\frac{w_{fk}h_{kn}}{\tilde{\lambda}_{fnk}}\right)
  \nonumber \\
  &\leq \sum_{k=1}^{K}\tilde{\lambda}_{fnk} \conv{d}_{\beta}\left(v_{fn} \;\middle|\;
  \frac{w_{fk}h_{kn}}{\tilde{\lambda}_{fnk}}\right)
  \label{eq:simplif} \\
  &\defequal \conv{G}_{fn}(\*W,\*H \mid \bWtilde, \bHtilde) \nonumber
  \, ,
\end{align}
where the coefficient $\tilde{\lambda}_{fnk}$ is defined as
$\tilde{\lambda}_{fnk} = \frac{\tilde{w}_{fk}\tilde{h}_{kn}}{\tilde{v}_{fn}}$ and we
denote $\bVtilde=\bWtilde\bHtilde$ with coefficients $\tilde{v}_{fn}$.
The concave term $\conc{d}_{\beta}$ is majorized using the tangent inequality
\begin{align*}
  \conc{d}_{\beta}&(v_{fn} \mid {[\*W\*H]}_{fn}) \\
  &\leq \conc{d}_{\beta}(v_{fn} \mid \tilde{v}_{fn})
  + \conc{d'}_{\beta}(v_{fn} \mid \tilde{v}_{fn})
  \left({[\*W\*H]}_{fn}-\tilde{v}_{fn}\right) \\
  &\defequal \conc{G}_{fn}(\*W,\*H | \bWtilde, \bHtilde)
  \, .
\end{align*}
Finally the overall auxiliary function $G$ is given by
\begin{align}
  \label{eq:jointG}
  G = \sum_{f=1}^{F}\sum_{n=1}^{N}\left[ \conv{G}_{fn} + \conc{G}_{fn} + \bar{d}_{\beta} \right]
  \, .
\end{align}
By construction, it satisfies the tight majorization properties
\begin{align*}
  &G(\*W,\*H \mid \bWtilde,\bHtilde) \geq D_{\beta}(\*V\mid\bWtilde\bHtilde) \\
  &G(\bWtilde,\bHtilde \mid \bWtilde,\bHtilde) = D_{\beta}(\*V\mid\bWtilde\bHtilde) \, ,
\end{align*}
which ensure the descent property in~\eqref{eq:desc_lemm}.
The expression of $G$ coincides with the joint auxiliary function derived
in~\cite[Table 2 in Section 4 and Appendix A]{Takahashi_N_2018_j-comput-optim-appl_unified_ugcamurnmf}
for a different purpose (see Section~\ref{sssec:cvg}).

We stress out that the auxiliary function $G$ is a tight joint majorization of
$D_{\beta}$.
This is in contrast with the BMM approach of Section~\ref{ssec:bmm} where two
separate auxiliary functions $G_{\*W}$ and $G_{\*H}$ are built for
$\*W \mapsto D_{\beta}(\*V|\*W\*H)$ and $\*H \mapsto D_{\beta}(\*V|\*W\*H)$, respectively.
A central difference in our JMM approach is the definition of the coefficients
$\{\tilde{\lambda}_{fnk}\}$ which depend on both current iterates
$\bWtilde$ and $\bHtilde$ and do not lead to a simplification of the term
$w_{fk} h_{kn}/\tilde{\lambda}_{fnk}$ in~\eqref{eq:simplif}.
This is in contrast with BMM, where, say, $\*W$ would be treated as a fixed
variable and the term $w_{fk} h_{kn}/\tilde{\lambda}_{fnk}$ simplifies to
$\tilde{v}_{fn}h_{kn}/\tilde{h}_{kn}$, allowing for closed-form minimization of
$G$.
Furthermore, the auxiliary function $G$ is not jointly convex, due to the
bilinear terms $w_{fk} h_{kn}$.
It is however bi-convex, i.e., convex w.r.t $\*W$ (resp., $\*H$) given $\*H$
(resp., $\*W$).

% ---------------------------------------------------------------------------- %

\subsection{Minimization step}
\label{ssec:min_step}

The auxiliary function $G$ does not appear to have a closed-form minimizer in
$\*W$ and $\*H$.
Neither is it convex, which makes it difficult to minimize globally.
While minimizing $G$ jointly in $\*W$ and $\*H$ is hard, we can still perform an
alternating minimization on the matrices $\*W$ and $\*H$ which results in the
following updates
\begin{equation}
  \label{eq:jnt_update}
  \begin{aligned}
    \*W &\; \longleftarrow \;
    \bWtilde.{\left(
      \frac{\frac{\*V}{{(\bWtilde\bHtilde)}^{.(2-\beta)}}{[\chi_{1,\beta}(\*H,\bHtilde)]}^{\top}}
           {{(\bWtilde\bHtilde)}^{.(\beta-1)}{[\chi_{2,\beta}(\*H,\bHtilde)]}^{\top}
           }\right)}^{.\gamma(\beta)} \\
    \*H &\; \longleftarrow \;
    \bHtilde.{\left(
      \frac{{[\chi_{1,\beta}(\*W,\bWtilde)]}^{\top}\frac{\*V}{{(\bWtilde\bHtilde)}^{.(2-\beta)}}}
           {{[\chi_{2,\beta}(\*W,\bWtilde)]}^{\top}{(\bWtilde\bHtilde)}^{.(\beta-1)}}\right)}^{.\gamma(\beta)}
    \, ,
  \end{aligned}
\end{equation}
where $\gamma(\beta)$ is defined as in~\eqref{eq:def_gamma} and
\begin{align*}
  \chi_{1,\beta}(\*H,\bHtilde) &=
  \begin{cases*}
    \frac{\bHtilde^{.(2-\beta)}}{\*H^{.(1-\beta)}} & if $\beta \leq 2$ \\
    \*H                                 & if $\beta > 2$ ,
  \end{cases*} \\
  \chi_{2,\beta}(\*H,\bHtilde) &=
  \begin{cases*}
    \*H                             & if $\beta < 1$ \\
    \frac{\*H^{.\beta}}{\bHtilde^{.(\beta-1)}} & if $\beta \geq 1$ .
  \end{cases*}
\end{align*}
The updates~\eqref{eq:jnt_update} are obtained by cancelling the partial
gradients of $G$.
Since $G$ is not convex, the alternating minimization may only lead to a
critical point instead of a global minimum.
Nevertheless, in order to decrease the loss function $D_{\beta}$, it is enough to
decrease $G$ thanks to the descent lemma~\eqref{eq:desc_lemm}.
This is easily achieved by initializing the updates~\eqref{eq:jnt_update} with
$(\bWtilde,\bHtilde)$, leading to Algorithm~\ref{algo:jmm_sketch}.

Note that, despite yielding a multiplicative update, the JMM updates given
in~\eqref{eq:jnt_update} are conceptually different from the BMM ones: the JMM
updates aim at minimizing $G$ given $\bWtilde, \bHtilde$ while the BMM updates
aim at minimizing $D_{\beta}(\*V|\*W \*H)$ w.r.t $\*H$ given $\*W$, or w.r.t $\*W$
given $\*H$.
Further comments and discussion are given in the next section.

\begin{algorithm}[t]
  \begin{algorithmic}[1]
    \renewcommand{\algorithmicrequire}{\textbf{Input:}}
    \renewcommand{\algorithmicensure}{\textbf{Output:}}
    \REQUIRE{Nonnegative matrix $\*V$ and initialization
    $(\*W_{\text{init}},\*H_{\text{init}})$}
    \ENSURE{Nonnegative matrices $\*W$ and $\*H$ such that $\*V \approx \*W\*H$}
    \STATE{Initialize $i$ to $1$}
    \STATE{Initialize $(\bWtilde_{i},\bHtilde_{i})$ to
      $(\*W_{\text{init}},\*H_{\text{init}})$}
    \REPEAT{}
    \STATE{$\bVtilde_{i} \leftarrow \bWtilde_{i}\bHtilde_{i}$}
    \STATE{Initialize $(\*W_{1},\*H_{1})$ to $(\bWtilde_{i},\bHtilde_{i})$}
    \FOR{$l=1 \dots L$} 
    \STATE{Update $\*W$ using~\eqref{eq:jnt_update}
      \[
      \*W_{l+1} \leftarrow
      \bWtilde_{i}.{\left(
        \frac{\frac{\*V}{{\bVtilde_{i}}^{.(2-\beta)}}
          {[\chi_{1,\beta}(\*H_{l},\bHtilde_{i})]}^{\top}}{{\bVtilde_{i}}^{.(\beta-1)}
          {[\chi_{2,\beta}(\*H_{l},\bHtilde_{i})]}^{\top}}\right)}^{.\gamma(\beta)}
      \]}
    \STATE{Update $\*H$ using~\eqref{eq:jnt_update}
      \[
      \*H_{l+1} \leftarrow
      \bHtilde_{i}.{\left(
      \frac{{[\chi_{1,\beta}(\*W_{l+1},\bWtilde_{i})]}^{\top}\frac{\*V}{{\bVtilde_{i}}^{.(2-\beta)}}}
           {{[\chi_{2,\beta}(\*W_{l+1},\bWtilde_{i})]}^{\top}{\bVtilde_{i}}^{.(\beta-1)}}\right)}^{.\gamma(\beta)}
      \]}
    \ENDFOR{} \\
    \STATE{$(\bWtilde_{i+1},\bHtilde_{i+1}) \leftarrow (\*W_{L+1},\*H_{L+1})$}
    \STATE{Increment $i$}
    \UNTIL{Convergence}
    \RETURN{$(\bWtilde_{i},\bHtilde_{i})$}
  \end{algorithmic}
  \caption{JMM~\label{algo:jmm_sketch}}
\end{algorithm}

% ---------------------------------------------------------------------------- %

\subsection{Discussion}
\label{ssec:discussion_jmm}

\subsubsection{Special cases}

For the values of notorious importance $0$, $1$ and $2$ of $\beta$, the
multiplicative updates~\eqref{eq:jnt_update} can be written in a simpler form as
some of the exponents cancel out and make the corresponding terms vanish.
Similar simplified updates are also available for the classic update
rules~\eqref{eq:alt_update}.
These simplified formulae are shown in Table~\ref{table:simpler_forms} for the
factor matrix $\*H$, where the matrix $\*1$ represents the matrix of dimension
$F \times N$ whose all entries equal $1$.

\begin{table}[t]
  \caption{Simplified updates of $\*H$ for $\beta=0,1,2$.}
  \begin{center}
    \setlength\tabcolsep{5.5pt}
    \resizebox{0.9\columnwidth}{!}{\begin{tabular}{cll}
      \toprule
      $\beta$ & \multicolumn{1}{c}{BMM} & \multicolumn{1}{c}{JMM} \\
      \midrule
      \midrule
      $0$ & $\*H \; \leftarrow\;\*H.{\left(\frac{\*W^{\top}\frac{\*V}{{(\*W\*H)}^{.2}}}
            {\*W^{\top}\frac{1}{\*W\*H}}\right)}^{.\frac{1}{2}}$
          & $\*H \; \leftarrow \;\bHtilde.{\left(\frac{{\left(\frac{\bWtilde^{.2}}{\*W}\right)}^{\top}
          \frac{\*V}{{(\bWtilde\bHtilde)}^{.2}}
            }
            {\*W^{\top}\frac{1}{\bWtilde\bHtilde}}\right)}^{.\frac{1}{2}}$ \\
      \midrule
      $1$ & $\*H \; \leftarrow \;\*H.\left(\frac{\*W^{\top}\frac{\*V}{\*W\*H}}{\*W^{\top}\*1}\right)$
          & $\*H \; \leftarrow \;\bHtilde.\left(\frac{\bWtilde^{\top}\frac{\*V}{\bWtilde\bHtilde}}
            {\*W^{\top}\*1}\right)$ \\
      \midrule
      $2$ & $\*H \; \leftarrow \;\*H.\left(\frac{\*W^{\top}\*V}{\*W^{\top}\*W\*H}\right)$
          & $\*H \; \leftarrow \;\bHtilde.\left(\frac{\*W^{\top}\*V}
            {{\left(\frac{\*W^{.2}}{\bWtilde}\right)}^{\top}\bWtilde\bHtilde}\right)$ \\
      \bottomrule
    \end{tabular}}
  \end{center}
  \label{table:simpler_forms}
\end{table}

% ~~~~~~~~~~~~~~~~~~~~~~~~~~~~~~~~~~~~~~~~~~~~~~~~~~~~~~~~~~~~~~~~~~~~~~~~~~~~ %

\subsubsection{Computational advantages of JMM}
\label{ssec:benef_jmm}

The new update rules~\eqref{eq:jnt_update} have a similar structure to the
classic multiplicative updates~\eqref{eq:alt_update} with a notable difference
regarding the matrices $\bWtilde$ and $\bHtilde$.
These matrices, named $\bWtilde_{i}$ and $\bHtilde_{i}$ in
Algorithm~\ref{algo:jmm_sketch}, remain constant in the sub-iterations and
allow for some computational savings w.r.t BMM when updating $\*W_{l}$ and
$\*H_{l}$.
For instance, the computation of $\bVtilde_{i}=\bWtilde_{i}\bHtilde_{i}$ is
performed only once per outer iteration in step 5 of
Algorithm~\ref{algo:jmm_sketch} whereas in Algorithm~\ref{algo:bmm_sketch} the
product $\*W\*H$ has to be computed at each update of $\*W_{l}$ (product
$\*W_{l}\bHcheck_{i}$ at step 6) and at each update of $\*W_{l}$ (product
$\bWcheck_{i+1}\*H_{l}$ at step 11).
Our proposed update hence saves here a matrix product per sub-iteration.
Similar computational savings can be obtained by storing for example, the
entry-wise ratio of $\*V$ and $\bVtilde$ in the update of JMM for $\beta=1$.

Table~\ref{table:nb_op} summarizes the computational savings (and extra divisions)
of JMM w.r.t BMM for the different values of $\beta$ at each iteration, i.e., for
one update of $\bWtilde_{i}$/$\bWcheck_{i}$ and one update of $\bHtilde_{i}$/$\bHcheck_{i}$.
For a fair comparison, we pick $L_{W}$ and $L_{H}$ equal to $L$.
The part multiplied by $L$ in the expressions in Table~\ref{table:nb_op}
corresponds to the computational savings obtained at each sub-iteration for
$\*W_{l}$ and $\*H_{l}$ while the remaining part corresponds to the extra cost
of computing matrices that are constant through the $L$ sub-iterations.
We especially emphasize the values $0$, $1$, and $2$ of $\beta$, which enjoy even
larger computational savings thanks to the simplifications shown in
Table~\ref{table:simpler_forms}.
Since these three cases are the most common in practice, the
update~\eqref{eq:jnt_update} brings a very welcome speedup compared
with~\eqref{eq:alt_update}. 
It turns out that the largest saving is in the case where $\beta$ is equal to $0$
or $1$, i.e., the IS and the KL divergences.
For other general values of $\beta$, the JMM updates incur extra divisions
in~\eqref{eq:jnt_update}, especially for $\beta$ in $]1,2[$, which mitigate the
computational savings.

%\begin{landscape}
\begin{table*}[t]
  \caption{Computational savings and extra divisions brought by JMM compared to
    BMM\@.
    The table reports the difference between the number of operations required
    by JMM and BMM per iteration of the outer loop, using $L_{W}=L_{H}=L$.
    Benefits of JMM are highlighted in bold font.}
  \begin{center}
    \setlength\tabcolsep{5.5pt}
    \resizebox{\columnwidth}{!}{\begin{tabular}{lccc}
      \toprule
               &      Multiplications &          Divisions
               & Additions \\
      \midrule
      $\beta=0$    & \boldmath$(2L-1)FNK + 2LFN$ & \boldmath$2(L-1)FN - L(FK+KN)$
               & \boldmath$(2L-1)FN(K-1)$ \\[0.3cm]
      $\beta=1$    & \boldmath$(4L-3)FNK$        & \boldmath$(2L-1)FN$
               & \boldmath$(4L-3)FNK - (L-1)(FN+FK+KN)$ \\[0.3cm]
      $\beta=2$    & \boldmath$(2L-1)FNK$        & $-L(FK+KN)$
               & \boldmath$(2L-1)FN(K-1)$ \\
      \midrule
      $\beta\in]1,2[$ & \boldmath$(2L-1)FNK -L(FK+KN)$ & $-L(2FN-FK-KN)$
                & \boldmath$(2L-1)FN(K-1)$ \\[0.3cm]
      $\beta>2$     & \boldmath$(2L-1)(FNK+FN)$      & $-L(FK+KN)$
                & \boldmath$(2L-1)FN(K-1)$ \\[0.3cm]
      $\beta<1$     & \boldmath$(2L-1)FNK$           & $-LK(FK+KN)-FN$
                & \boldmath$(2L-1)FN(K-1)$ \\
      \bottomrule
      \end{tabular}}
  \end{center}
  \label{table:nb_op}
\end{table*}
%\end{landscape}

% ~~~~~~~~~~~~~~~~~~~~~~~~~~~~~~~~~~~~~~~~~~~~~~~~~~~~~~~~~~~~~~~~~~~~~~~~~~~~ %

\subsubsection{Failure of heuristic updates}

A common heuristic in BMM consists in setting $\gamma(\beta)$ to $1$ in the
update~\eqref{eq:alt_update}, even for values of $\beta$ outside $[1,2]$.
Indeed, it has been empirically observed that this leads to an equally good
factorization while decreasing the number of iterations to reach convergence.
For values of $\beta$ in $[0,1]$, the authors
in~\cite{Fevotte_C_2011_j-neural-comput_algorithms_nmfbd} have proved that this
heuristic corresponds to a Majorization-Equalization scheme which produces
larger steps than the MM method.
Nevertheless, deriving theoretical support for this heuristic for other values
of $\beta$ is still an open problem.
Setting $\gamma(\beta) = 1$ for all $\beta$ did not lead to similar findings for JMM\@.
While we did observe that the objective function decreases at every iteration
under the heuristic, worse solutions were obtained (i.e., corresponding to
higher values of the objective function in general).
This might be due to the non-convexity of the JMM auxiliary function $G$, in
which case the Majorization-Equalization principle makes less sense.

% ~~~~~~~~~~~~~~~~~~~~~~~~~~~~~~~~~~~~~~~~~~~~~~~~~~~~~~~~~~~~~~~~~~~~~~~~~~~~ %

\subsubsection{Convergence of the iterates of JMM}
\label{sssec:cvg}

In standard NMF practice, BMM is used with $L_{W}=L_{H}=1$ sub-iterations.
The convergence of the iterates of BMM in this setting can be proven for
slightly modified NMF problems that essentially ensure the coercivity of the
objective function on its domain (loosely speaking, $f({\bm \theta}) \rightarrow \infty$ whenever
$\| {\bm \theta} \| \rightarrow \infty$).
In~\cite{Zhao_R_2018_j-ieee-trans-sig-proc_unified_camuarnmf} this is ensured by
augmenting the objective function with an $\ell_{1}$ regularization term on $\*W$
and $\*H$.
Then the convergence of the iterates can be invoked using the Block Successive
Upper-bound Minimization (BSUM) framework of~\cite{Razaviyayn_M_2013_j-siam-j-optim_unified_cabcmmno}.
In~\cite{Takahashi_N_2018_j-comput-optim-appl_unified_ugcamurnmf}, coercivity is
ensured by replacing the nonnegativity constraint with a strict positivity
constraint $\*W,\*H \ge \varepsilon$.
Then the convergence of the iterates can be invoked using Zangwill's
convergence theorem by astutely reformulating BMM (with $L_{W}=L_{H}=1$) as
follows: $\forall l \ge 1$,
\begin{align}
  \*W_{l+1} &= \argmin_{ \*W \geq \varepsilon} \ G(\*W,\*H_{l} \mid \*W_{l},\*H_{l}) \label{eq:Wll} \\
  \*H_{l+1} &= \argmin_{ \*H \geq \varepsilon} \ G(\*W_{l+1},\*H \mid  {\*W_{l+1}},\*H_{l}) \label{eq:Hll}.
\end{align}
This is how the joint auxiliary function $G$ given by~\eqref{eq:jointG} is
introduced in~\cite{Takahashi_N_2018_j-comput-optim-appl_unified_ugcamurnmf},
namely as a convenient way to derive BMM from a unique function of four
variables, rather than using separate auxiliary functions (of two variables)
for each sub-problem.
The fundamental difference between BMM and our novel JMM approach is that the
second occurrence of $\*W_{l+1}$ in~\eqref{eq:Hll} is left unchanged in JMM
(i.e., it remains $\*W_{l}$). This seemingly trivial change has significant
computational implications in practice (and can only be justified by the joint
MM framework that we introduced).

The proofs of convergence in~\cite{Zhao_R_2018_j-ieee-trans-sig-proc_unified_camuarnmf,
  Takahashi_N_2018_j-comput-optim-appl_unified_ugcamurnmf} heavily rely on the
block-structure of BMM and the strict convexity of the auxiliary
functions~\eqref{eq:Wll} w.r.t. $\*W$ and~\eqref{eq:Hll} w.r.t. $\*H$.
Single-block MM algorithm also requires being able to find a global minimizer
of the auxiliary function~\cite{Lange_K_2013_book_optimization}.
Without this requirement, it is not possible to determine in general whether
the limit points of the sequence of iterates are critical points of the
initial objective function.
Unfortunately, the auxiliary function $G$, which lies at the heart of JMM,
is not jointly convex w.r.t.\ its first two variables.
With $L$ sufficiently large, the alternating minimization steps 7 and 8 in
Algorithm~\ref{algo:jmm_sketch} only guarantee that we find a critical point
of $G$ at every iteration, which is not necessarily a global minimum.
Hence, proving the convergence of the iterates of our JMM method (or more
generally of MM algorithms with non-convex auxiliary functions) is a difficult
problem that is left for future work.
Remember however that the convergence of the objective function is guaranteed
by design (for any value of $L$).

% ============================================================================ %

\section{Experimental Results}
\label{sec:simul}

We provide in this section extensive numerical comparisons of BMM and JMM using
various datasets (face images, audio spectrograms, song play-counts,
hyperspectral images) with diverse dimensions.

% ---------------------------------------------------------------------------- %

\subsection{Set-up}

Our implementation for JMM follows Algorithm~\ref{algo:jmm_sketch} while the one
for BMM follows Algorithm~\ref{algo:bmm_sketch}.
Some additional practical considerations are detailed below.

\subsubsection{Influence of the number of sub-iterations}

\begin{figure}[!t]
  \centering
  \includegraphics[width=\linewidth]{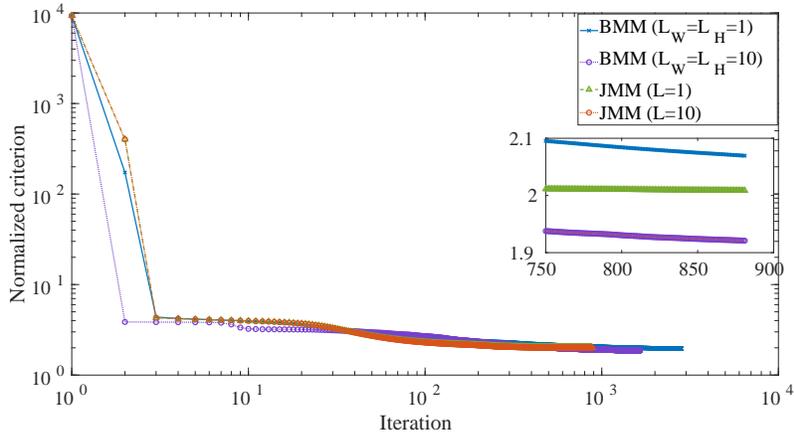}                  
  \caption{Impact of the number of sub-iterations in the minimization step of
    Algorithm~\ref{algo:jmm_sketch}.
    The box is a zoom-in on iterations $750$ to $881$.
    In the latter, BMM ($L_{W}=L_{H}=10$) and JMM ($L=10$) are overlapping.}
  ~\label{fig:comp_vs_ssiter_nmf}
\end{figure}

Algorithm~\ref{algo:jmm_sketch} dictates that we update $\*W$ and $\*H$ several
times in the inner loop to fully minimize
$(\*W,\*H) \mapsto G(\*W,\*H | \bWtilde,\bHtilde)$
(without updating or recomputing the tilde matrices).
Nevertheless, this does not turn out particularly advantageous in practice.
Indeed, in all our simulations, we observe that only the first iteration
of~\eqref{eq:jnt_update} results in a significant decrease of $G$ and that the
additional sub-iterations yield only negligible improvement.
This is illustrated in Figure~\ref{fig:comp_vs_ssiter_nmf} which displays the
objective function values obtained with JMM and BMM (as a function of the outer
iterations) for different numbers of sub-iterations $L$, $L_{W}$, and $L_{H}$,
using the Olivetti face dataset (see Section~\ref{ssec:fac_img}) and $\beta=1$.
Figure~\ref{fig:comp_vs_ssiter_nmf} shows that the plots for JMM with $L=1$ and
$L=10$ sub-iterations are nearly overlapping.

As such, like in traditional NMF practice, in the following we only use one
sub-iteration of $\*W$ and $\*H$ in our implementations of JMM and BMM\@.
This means we set $L=1$ in Algorithm~\ref{algo:jmm_sketch} and $L_{W}=L_{H}=1$
in Algorithm~\ref{algo:bmm_sketch}.
By doing so, the BMM and JMM updates of $\*W$ coincide and only the update of
$\*H$ changes (with still a significant gain in performance).
Note that we have observed empirically that inverting the order of the updates
in~\eqref{eq:jnt_update} does not have any impact on the number of iterations
before convergence nor on the quality of the obtained solutions.

% ~~~~~~~~~~~~~~~~~~~~~~~~~~~~~~~~~~~~~~~~~~~~~~~~~~~~~~~~~~~~~~~~~~~~~~~~~~~~ %

\subsubsection{Initialization and stopping criterion}

The initializations $(\*W_{\text{init}},\*H_{\text{init}})$ for BMM and JMM are
drawn randomly according to a half-normal distribution.
Our stopping criterion for both algorithms is based on the relative decrease in
the objective function $D_{\beta}$.
More precisely, the algorithms are stopped when
\begin{equation*}
  \frac{D_{\beta}(\*V \mid \bar{\*W}\bar{\*H})-D_{\beta}(\*V \mid \*W\*H)}{D_{\beta}(\*V \mid \*W\*H)}
  \leq \epsilon \; ,
\end{equation*}
where $\epsilon$ is a tolerance set to $10^{-5}$, $\*W$ and $\*H$ are the current
outer-loop iterates while $\bar{\*W}$ and $\bar{\*H}$ are the previous ones
(either $\bWcheck$ and $\bHcheck$ for BMM or $\bWtilde$ and $\bHtilde$ for JMM).
Furthermore, in order to remove the scaling ambiguity inherent to NMF, we
normalize $\*W$ and $\*H$ at the end of each outer-loop iteration for both BMM
and JMM\@.
The normalization consists in dividing each column of $\*W$ by its $\ell_{2}$ norm and
scaling the rows of $\*H$ accordingly.

% ~~~~~~~~~~~~~~~~~~~~~~~~~~~~~~~~~~~~~~~~~~~~~~~~~~~~~~~~~~~~~~~~~~~~~~~~~~~~ %

\subsubsection{Handling zero values and numerical stability}

The $\beta$-divergence $D_{\beta}(x|y)$ may not be defined when either $x$ or $y$ takes
zero values.
This is for instance the case when $\beta$ is set to $0$ or $1$ due to the quotient
and the logarithm that appear.
As such, we recommend minimizing $D_{\beta}(\*V+\kappa\*1|\*W\*H+\kappa\*1)$ with a small
constant $\kappa$ instead of $D_{\beta}(\*V|\*W\*H)$ for numerical stability.
This first simply amounts to replacing $\*V$ by $\*V + \kappa\*1$ in the previous
derivations.
Then, by treating $\kappa$ as a ${(K + 1)}^{\text{th}}$ constant component like
in~\cite{Fevotte_C_2015_j-ieee-trans-img-proc_nonlinear_hurnmf}, we may simply
replace the product $\bWtilde\bHtilde$ (resp. $\*W\*H$) by $\bWtilde\bHtilde + \kappa\*1$
(resp. $\*W\*H + \kappa\*1$) in~\eqref{eq:jnt_update} (resp.~\eqref{eq:alt_update}).

% ~~~~~~~~~~~~~~~~~~~~~~~~~~~~~~~~~~~~~~~~~~~~~~~~~~~~~~~~~~~~~~~~~~~~~~~~~~~~ %

\subsubsection{Simulation environment}

All the simulations have been conducted in Matlab 2020a running on an Intel
i7-8650U CPU with a clock cycle of 1.90GHz shipped with 16GB of
memory.\footnote{Matlab code is available at
\href{https://arthurmarmin.github.io/research.html}{https://arthurmarmin.github.io/research.html}.}
In each of the following experimental scenarii, we compare the factorization
obtained by JMM and BMM from $25$ different initializations
$(\*W_{\text{init}},\*H_{\text{init}})$.

% ~~~~~~~~~~~~~~~~~~~~~~~~~~~~~~~~~~~~~~~~~~~~~~~~~~~~~~~~~~~~~~~~~~~~~~~~~~~~ %

\subsubsection{Performance evaluation}

We compare BMM and JMM both in terms of computational efficiency (CPU time) 
and quality of the returned solutions.
To assess the latter, we use the value of the normalized objective function
$\frac{D_{\beta}(\*V|\widehat{\*W}\widehat{\*H})}{FN}$ at the solution
$(\widehat{\*W},\widehat{\*H})$ returned by the NMF algorithms.
We also consider the KKT
residuals~\cite{Fevotte_C_2011_j-neural-comput_algorithms_nmfbd} to measure the
distance to the first order optimality conditions and attest whether the
algorithms reached a critical point of the criterion $D_{\beta}$.
The KKT residuals are expressed by
\begin{align*}
  \text{res}(\*W)
  &= \frac{\norm{\min\{\*W, [{(\*W\*H)}^{.(\beta-2)}.(\*W\*H-\*V)]\*H^{\top}\}}_{1}}{FK}
  \\
  \text{res}(\*H)
  &= \frac{\norm{\min\{\*H, \*W^{\top}[{(\*W\*H)}^{.(\beta-2)}.(\*W\*H-\*V)]\}}_{1}}{KN}
  \, .
\end{align*}

% ---------------------------------------------------------------------------- %

\subsection{Factorization of face images}
\label{ssec:fac_img}

In the context of image processing, NMF can be used to learn part-based features
from a collection of images~\cite{Lee_D_1999_j-nature_learning_ponmf}.
The columns of the matrix $\*V$ correspond to the vectorization of the different
images.
Besides, the factor $\*W$ represents the dictionary of image features, and the
matrix $\*H$ contains the activation encodings.

We compare the BMM and JMM methods on a face images dataset, the Olivetti
dataset from AT\&T Laboratories
Cambridge~\cite{Samaria_F_1994_p-wacv_parametrisation_smhfi}, which contains
$400$ greyscale images of faces with dimensions $64 \times 64$.
The corresponding matrix $\*V$ thereby has dimensions $4096 \times 400$.
We set $K$ to $10$.
We consider the values $0$, $1$, and $2$ for the parameter $\beta$, which
correspond respectively to IS divergence, KL divergence, and squared Euclidean
distance (the latter two being the most common in image processing).

\begin{figure}[!t]
  \centering
  \begin{subfigure}[b]{\columnwidth}
    \includegraphics[width=\linewidth]{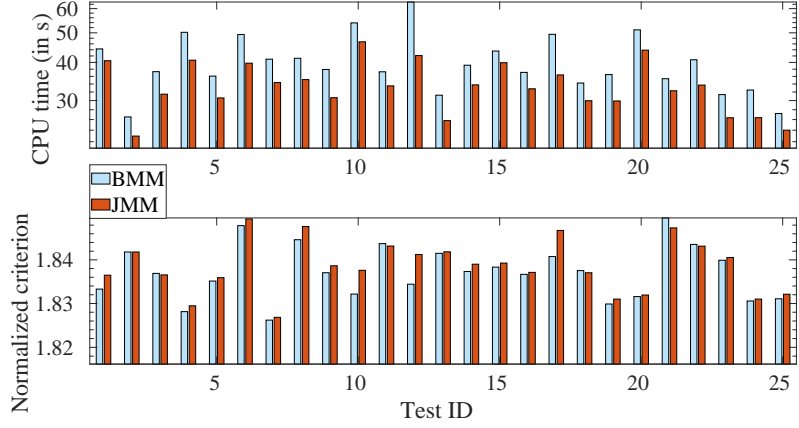}
    \caption{Using KL divergence ($\beta=1$).}
  \end{subfigure}
  \begin{subfigure}[b]{\columnwidth}
    \includegraphics[width=\linewidth]{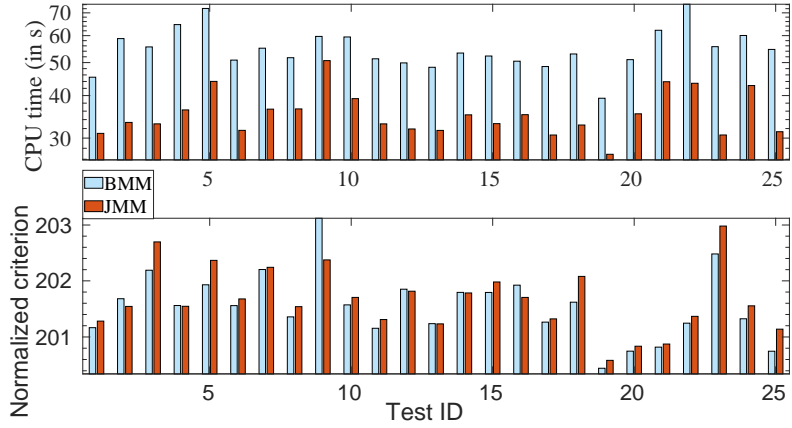}
    \caption{Using quadratic loss ($\beta=2$).}
  \end{subfigure}     
  \caption{Comparative performance using the Olivetti dataset ($K=10$).}
  ~\label{fig:oliv_results_K=10}
\end{figure}

Figure~\ref{fig:oliv_results_K=10} shows the computation times for $\beta=2$ and
$\beta=1$, as well as the values of the normalized objective function produced by
both BMM and JMM, for the $25$ runs.
The same random initialization is used by both methods.
Note that we use a logarithmic scale for the CPU time and that the y-axis for
the objective function does not start at zero.
We observe that JMM is always faster than BMM while yielding solutions with a
similar quality in terms of the objective $D_{\beta}$.
The corresponding average CPU time as well as the resulting acceleration ratios
are given in Table~\ref{table:overall_stat}.
We notice that the computational saving is higher when using the IS divergence,
which confirms our analysis from Section~\ref{ssec:benef_jmm}.
Remark that we measure here the global CPU time and not the time per iteration:
since the auxiliary function is different for BMM and JMM, the trajectory in the
parameter space is also different and thus the two algorithms do not require the
same number of iterations before convergence.
Nevertheless, we observe that the number of iterations for both algorithms has a
similar order of magnitude.
Since the iterations of JMM are cheaper, its global CPU time is lower.
This remark also holds for higher dimensional dataset such as the one
in Section~\ref{ssec:tasteprof}.

Note that while MM algorithms are prevalent for $\beta$-NMF in general, many other
algorithms have been designed for the specific case $\beta =2$
(quadratic loss)~\cite{Gillis_N_2020_book_nonnegative_mf}.
In that case, some algorithms are notoriously more efficient than BMM,
see, e.g.,~\cite[Chapter 8.2]{Gillis_N_2020_book_nonnegative_mf}.
Though JMM improves on BMM, it may not compete with these other algorithms.
Still, JMM, like BMM, is free of hyper-parameters and very easy to use, and
remains a very convenient option for the general practitioner.

Finally, we computed the KKT residuals for $\widehat{\*W}$ and $\widehat{\*H}$
returned by JMM and BMM, for all runs and considered values of $\beta$.
We observed that they range from $1.10^{-5}$ to $1.10^{-1}$, indicating that
both methods converge to a critical point of $D_{\beta}$.
As a matter of fact, an additional significant observation is that in all cases
here, both JMM and BMM return the same solution $(\hat{\*W},\hat{\*H})$ up to
permutation of their columns and up to some round-off errors.
Nevertheless, the trajectory of the iterates may differ.

% ---------------------------------------------------------------------------- %

\subsection{Factorization of a spectrogram}

We now consider NMF of audio magnitude spectrograms.
In this context the factor $\*W$ contains elementary audio spectra with temporal
activations given by
$\*H$~\cite{Smaragdis_P_2014_j-ieee-sig-proc-mag_static_dssunmfuv}. 

We generate the spectrogram of an excerpt from the original recording of the
song ``Four on Six'' by Wes Montgomery.
The signal is $50$-seconds long with a sampling rate of \SI{44.1}{\kilo\hertz}.
The spectrogram is computed with a Hamming window of length $2048$ (46ms) and an
overlap of $50\%$.
This results in a data matrix $\*V$ of dimensions $1025 \times 2152$.

\begin{figure}[!t]
  \centering
  \includegraphics[width=\linewidth]{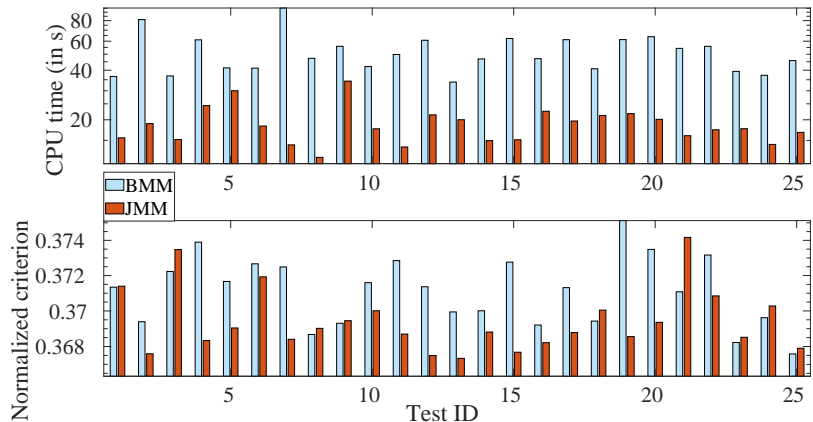}
  \caption{Comparative performance with a spectrogram
    \\\hspace{\textwidth}($K=10,\beta=0$).}
  ~\label{fig:spectro_res_K=10}
\end{figure}

In this section, we set $\beta=0$, which is a common value in audio signal
processing~\cite{Fevotte_C_2009_j-neural-comput_nonnegative_mfisdama}, and set
$K$ to $10$.
The results obtained with BMM and JMM are shown on
Figure~\ref{fig:spectro_res_K=10}.
The average computation time for BMM is 291s while the one for JMM is 41s, which
yields an average acceleration of $86\%$.
We observe a very substantial benefit of JMM over BMM in terms of CPU time in this
case.
This confirms our conclusion from Section~\ref{ssec:benef_jmm} that JMM reaches
its highest potential with NMF based on the IS divergence.
We observe here again that JMM and BMM return the same solution
$(\hat{\*W},\hat{\*H})$ up to permutation of their columns.

% ---------------------------------------------------------------------------- %

\subsection{Factorization of song play-counts}
\label{ssec:tasteprof}

NMF may be used in recommendation systems based on implicit user data.
In this case, the matrix $\*V$ contains information about the interactions of
users with a collection of items.
The factor $\*W$ may extract user preferences while $\*H$ represents item
attributes, see, e.g.,~\cite{Hu_Y_2008_p-ieee-icdm_collaborative_fifd}.

We here consider the TasteProfile
dataset~\cite{Bertin-Mahieux_T_2011_book_million_sd} which contains
counts of songs played by users (e.g., of a music streaming service).
Similarly to~\cite{Gouvert_O_2020_p-icml_ordinal_nmdr} and many other papers using
this dataset, we apply a preprocessing to the original data to keep only users
and songs with more than a given number of interactions (here set to 20).
This results in a large and sparse data matrix $\*V$ of dimensions
$16\,301 \times 12\,118$ with about $0.6\%$ nonzero values.
We set $\beta$ to $1$ (a common choice in recommender systems, because its
corresponds to the log-likelihood of a Poisson model that is natural for count
data), and $K=50$.

\begin{figure}[!t]
  \centering
  \includegraphics[width=\linewidth]{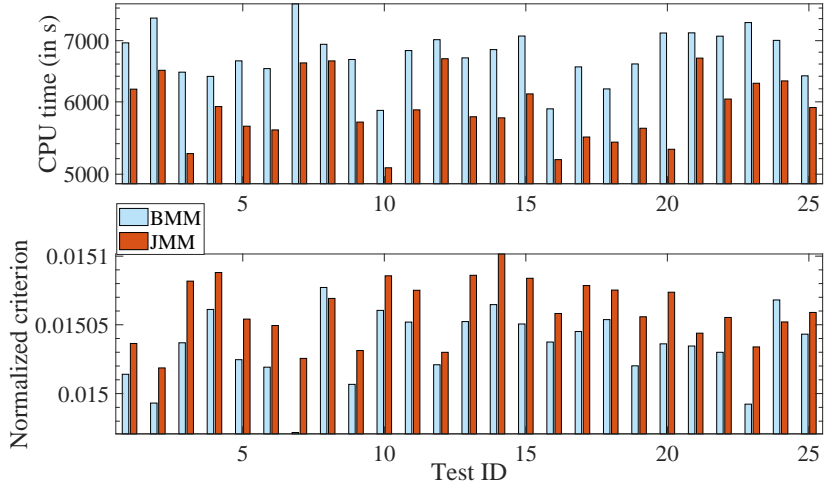}                  
  \caption{Comparative performance with TasteProfile
    \\\hspace{\textwidth}($K=50,\beta=1$).}
  ~\label{fig:tasteprof_res}
\end{figure}

Comparative results are displayed in Figure~\ref{fig:tasteprof_res}.
We observe that on average, JMM is $13\%$ faster with an average CPU time of
1 hour 38 minutes whereas BMM's average time is equal to 1 hour 58 minutes.
Finally, like in previous scenarii, JMM and BMM return the same solution
$(\hat{\*W},\hat{\*H})$ up to a permutation of their columns.

Similarly to our observations of Section~\ref{ssec:fac_img}, we observe that,
while BMM has generally a shorter trajectory than BMM, the number of iterations
of both methods has the same order of magnitude.
Since the cost of BMM per iteration is higher than JMM, this results in a
higher overall CPU time.

% ---------------------------------------------------------------------------- %

\subsection{Factorization of hyperspectral images}

A hyperspectral image is a multi-band image that can be represented by a
nonnegative matrix: each row represents a spectral band while each column
represents a pixel of the image.
Applying NMF to such matrix data allows extracting a collection of individual
spectra representing the different materials arranged in the matrix $\*W$, as
well as their relative proportions given by the matrix $\*H$, see,
e.g.,~\cite{BioucasDias_J_2012_j-ieee-j-sel-top-appl-earth-obs-rem-sens_hyperspectral_uogssrba}.

We consider a hyperspectral image acquired over Moffett Field in 1997 by the
Airborne Visible Infrared Imaging Spectrometer~\cite{Aviris_database}.
The image contains $50 \times 50$ pixels over $189$ spectral bands, which leads to a
matrix $\*V$ of dimensions $189 \times 2500$.
We consider $\beta=2$ and $\beta=1.5$.
The latter value in particular was shown to be an interesting trade-off between
Poisson ($\beta=1$) and additive Gaussian ($\beta=2$) assumptions for predicting missing
values in incomplete versions of this dataset,
see~\cite{Fevotte_C_2015_j-ieee-trans-img-proc_nonlinear_hurnmf}.
The factorization rank $K$ is set to $3$, a standard choice with this dataset
(extraction of vegetation, soil and water).

\begin{figure}[!t]
  \centering
  \begin{subfigure}[b]{\columnwidth}
    \includegraphics[width=\linewidth]{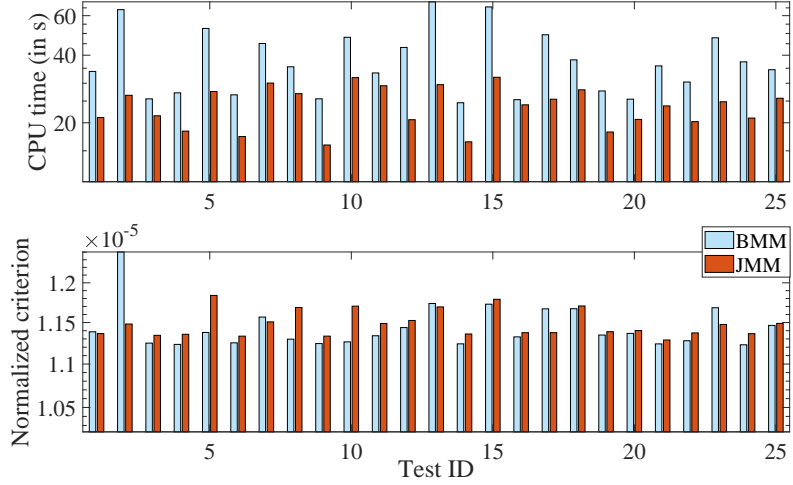}
    \caption{$\beta=2$.}
  \end{subfigure}
  \begin{subfigure}[b]{\columnwidth}
    \includegraphics[width=\linewidth]{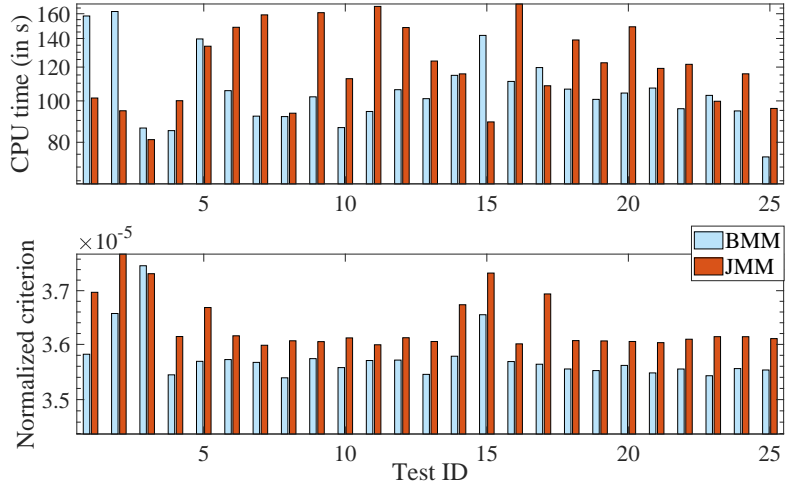}
    \caption{$\beta=1.5$.}
  \end{subfigure}     
  \caption{Comparative performance with Moffet hyperspectral image ($K=3$).}
  ~\label{fig:hyperimg_moffet}
\end{figure}

Comparative results are given in Figure~\ref{fig:hyperimg_moffet}.
On the top figure corresponding to $\beta=2$, we observe that JMM is
faster than BMM with an acceleration of $31\%$ on average. 
On the other hand, for $\beta=1.5$, we observe that BMM is faster in general (though
not always).
This confirms again our analysis in Section~\ref{ssec:benef_jmm}: when $\beta$ is
different from $0$, $1$, and $2$, the benefit of JMM may be counterbalanced by
the additional divisions in the general formulae~\eqref{eq:jnt_update}.

% ---------------------------------------------------------------------------- %

\subsection{Summary of the experiments}

Table~\ref{table:overall_stat} summarizes the average and the standard
deviation of the CPU times for all the datasets considered.
It can be observed that JMM yields a significant speed-up for the widely used
settings when $\beta$ equals to $0$, $1$, or $2$.
This acceleration is observed on small, medium and large size datasets.
The corresponding $95\%$ confidence intervals are given in
Table~\ref{table:confid_int}.

\begin{table}[t]
  \caption{Summary for the statistics on CPU times and the acceleration
    in all the tested dataset.
    The displayed values are the averages while the standard deviations are
    indicated within parenthesis.}
    \begin{center}
      \setlength\tabcolsep{5.5pt}
      \begin{tabular}{lccc}
        \toprule
        & BMM & JMM & Acc. \\
        \midrule
        Olivetti \hfill ($\beta=2$)     &   55s    (8s) &   36s    (6s) & 35\% \\
        Olivetti \hfill ($\beta=1$)     &   40s    (9s) &   34s    (6s) & 16\% \\
        Olivetti \hfill ($\beta=0$)     &  229s   (32s) &   63s    (8s) & 72\% \\
        Spectrogram \hfill ($\beta=0$)  &  291s   (90s) &   41s    (8s) & 86\% \\
        TasteProfile \hfill ($\beta=1$) & 6776s  (439s) & 5909s  (492s) & 13\% \\
        Moffet \hfill ($\beta=2$)       &   39s   (13s) &   24s    (5s) & 35\% \\
        Moffet \hfill ($\beta=1.5$)     &  107s   (22s) &  123s   (26s) & -18\% \\
        \bottomrule
      \end{tabular}
    \end{center}
    \label{table:overall_stat}
\end{table}

\begin{table}[t]
  \caption{The $95\%$ confidence intervals for the CPU times of BMM and JMM
    (in seconds).}
    \begin{center}
      \setlength\tabcolsep{5.5pt}
      \begin{tabular}{lcc}
        \toprule
        & BMM & JMM \\
        \midrule
        Olivetti \hfill ($\beta=2$)     &   $[52,  58]$ &   $[34,  38]$ \\
        Olivetti \hfill ($\beta=1$)     &   $[36,  44]$ &   $[32,  36]$ \\
        Olivetti \hfill ($\beta=0$)     &  $[216, 242]$ &   $[60,  66]$ \\
        Spectrogram \hfill ($\beta=0$)  &  $[254, 328]$ &   $[38,  43]$ \\
        TasteProfile \hfill ($\beta=1$) & $[6595,6957]$ & $[5706,6112]$ \\
        Moffet \hfill ($\beta=2$)       &   $[34,  44]$ &   $[22,  26]$ \\
        Moffet \hfill ($\beta=1.5$)     &   $[98, 116]$ &  $[112, 133]$ \\
        \bottomrule
      \end{tabular}
    \end{center}
    \label{table:confid_int}
\end{table}

% ============================================================================ %

\section{Conclusion}
\label{sec:concl}

In this paper, we have presented a joint MM method for NMF with the
$\beta$-divergence.
Our algorithm relies on the alternating minimization of a non-convex auxiliary
function and leads to new multiplicative updates of the factor matrices.
These new updates are variants of the classic multiplicative updates and are
equally simple to implement.
They can lead to a significant computational speedup,
especially for the Itakura-Saito and Kullback Leibler divergences, and the
quadratic loss.
Fortunately, the three later cases are the three most common cases of NMF with
the $\beta$-divergence.
The new updates guarantee the descent property for the objective function.
Moreover, we have observed experimentally that the estimated factors are of the
same quality as the ones returned by the classic block MM scheme.
As a matter of fact, we have noted that both algorithms usually return the same
solutions up to column permutation.
The computational efficiency of the proposed updates has been demonstrated on
datasets with diverse characteristics.
In future work, we intend to study the convergence of the iterates of JMM\@.
This is a challenging topic because of the non-convexity of the majorizer which
underpins our approach.
Such a difficulty explains the scarcity of results in the literature on the
convergence of MM algorithms with non-convex auxiliary functions.
Another promising topic would consist in designing stochastic versions of JMM
for the factorization of massive datasets~\cite{Mensch_A_2018j-ieee-trans-sig-proc_stochastic_sfhm,
  Pu_W_2022_j-ieee-trans-sig-proc_stochastic_mdlrtdnl}.

% =========================================================================== %

\section*{Acknowledgment}
The authors acknowledge R{\'e}mi Flamary, J{\'e}r{\^o}me Idier, Paul Magron and
Emmanuel Soubies for discussions related to this work.

% =========================================================================== %

\bibliographystyle{elsarticle-num}
\bibliography{abbr,mybiblio}

% =========================================================================== %

\end{document}